\def\year{2022}\relax

\documentclass[letterpaper]{article} 
\usepackage{aaai22}  
\usepackage{times}  
\usepackage{helvet}  
\usepackage{courier}  
\usepackage[hyphens]{url}  
\usepackage{graphicx} 
\usepackage{natbib}  
\usepackage{caption} 
\DeclareCaptionStyle{ruled}{labelfont=normalfont,labelsep=colon,strut=off} 
\usepackage{makecell}
\usepackage{latexsym}
\usepackage{amsmath}
\usepackage{amsfonts}
\usepackage{booktabs}
\usepackage{hhline}
\usepackage{color}
\usepackage{bm}
\usepackage{subfigure}

\begin{document}
%
\title{KGR$^{4}$: Retrieval, Retrospect, Refine and Rethink for Commonsense Generation}
\author{
Xin Liu\textsuperscript{1}~~~Dayiheng Liu\textsuperscript{2}~~~Baosong Yang\textsuperscript{2}~~~Haibo Zhang\textsuperscript{2}~~~Junwei Ding\textsuperscript{2}\\
Wenqing Yao\textsuperscript{2}~~~Weihua Luo\textsuperscript{2}~~~Haiying Zhang\textsuperscript{1}~~~Jinsong Su\textsuperscript{1,\footnotemark[1]}\\
}
\affiliations {
    \textsuperscript{1}School of Informatics, Xiamen University~~~
    \textsuperscript{2}Alibaba Group\\
    \texttt{\small{liuxin@stu.xmu.edu.cn}} \\
  \texttt{\small{\{liudayiheng.ldyh, yangbaosong.ybs, zhanhui.zhb, djw99219,\\ wenqing.ywq, weihua.luowh\}@alibaba-inc.com}}\\
  \texttt{\small{\{zhang2002,jssu\}@xmu.edu.cn}}
}
\maketitle
\begin{abstract}
\begin{quote}
Generative commonsense reasoning requires machines to generate sentences describing an everyday scenario given several concepts, which has attracted much attention recently. 
However, existing models cannot perform as well as humans, since sentences they produce are often implausible and grammatically incorrect.
In this paper, inspired by the process of humans creating sentences, we propose a novel \textbf{K}nowledge-enhanced Commonsense \textbf{G}eneration framework, termed KGR$^\mathfrak{4}$, consisting of four stages: \textbf{R}etrieval, \textbf{R}etrospect, \textbf{R}efine, \textbf{R}ethink.
Under this framework, we first perform retrieval to search for relevant sentences from external corpus as the prototypes. Then, we train the generator that either edits or copies these prototypes to generate candidate sentences, of which potential errors will be fixed by an autoencoder-based refiner. 
Finally, we select the output sentence from candidate sentences produced by generators with different hyper-parameters.
Experimental results and in-depth analysis on the CommonGen benchmark strongly demonstrate the effectiveness of our framework. Particularly, KGR$^\mathfrak{4}$ obtains 33.56 \texttt{SPICE} points in the official leaderboard, outperforming the previously-reported best result by 2.49 \texttt{SPICE} points and achieving state-of-the-art performance.\footnote{Till our submission time, 18 May. 2021. We refer readers to \url{https://inklab.usc.edu/CommonGen/leaderboard.html} for the latest results. We release the code at https://github.com/DeepLearnXMU/KGR-4.}


\end{quote}
\end{abstract}

\renewcommand{\thefootnote}{\fnsymbol{footnote}}
\footnotetext[1]{Jinsong Su is the corresponding author. This work was done when Xin Liu was interning at DAMO Academy, Alibaba Group.}

\section{Introduction}
Recently, integrating commonsense knowledge into artificial intelligence models has become increasingly attractive to researchers.
To assess the ability of these models in understanding the commonsense knowledge from our daily life, various tasks \citep{DBLP:conf/emnlp/ZellersBSC18, talmor-etal-2019-commonsenseqa, DBLP:conf/acl/ZellersHBFC19, lin-etal-2020-commongen} for commonsense reasoning have been proposed. 
Typically, SWAG \citep{DBLP:conf/emnlp/ZellersBSC18}, HellaSWAG \citep{DBLP:conf/acl/ZellersHBFC19}, and CommonsenseQA \citep{talmor-etal-2019-commonsenseqa} are designed to infer an upcoming event by selecting one of the listed choices.
However, none of them focus on commonsense reasoning in a generative manner, which is considered as a basic ability of human beings \citep{moore2013development}.
To deal with this issue, \citet{lin-etal-2020-commongen} explore CommonGen, which requires the model to produce the sentence describing the daily life scenario given some concepts.
As shown in the first two lines of Table \ref{data_case}, given concepts \{\emph{hand, sink, wash, soap}\}, this task aims to generate a coherent sentence covering all of them, e.g. ``\emph{The girl uses soap to wash her hands at the sink}''. 
Compared with previous tasks, CommonGen is able to better evaluate the generative commonsense reasoning ability of each model, thus attracting much attention recently.

However, CommonGen remains a difficult task due to the following challenges: 1) Generated sentences should be consistent with commonsense knowledge; 2) The model is required to possess the compositional generalization ability, so that it can deal with unseen combinations of concepts. 
From Table \ref{data_case}, we can observe that the sentences generated by commonly-used pretrained models are either not in line with commonsense knowledge (i.e., ``\emph{a sink of soaps}'' produced by BERT-Gen \citep{pmlr-v119-bao20a}) or suffer from the repetition problem (i.e., ``\emph{in a sink a sink}'' generated by BART \citep{lewis-etal-2020-bart}).

\renewcommand\arraystretch{1.0}
\begin{table}[t]
\centering
\scalebox{0.9}{
\begin{tabular}{ll}
\bottomrule
\multicolumn{2}{l}{\textbf{Concepts}: \{\emph{hand, sink, wash, soap}\}}                                                                                             \\
\multicolumn{2}{l}{\textbf{Outputs:} The girl uses \emph{soap} to \emph{wash} her \emph{hands} at the \emph{sink}.}                                                                       \\ \hline
\multicolumn{2}{l}{\textbf{GPT-2:} \emph{hands} \emph{washing} \emph{soap} on the \emph{sink}.}                                                                                           \\
\multicolumn{2}{l}{\textbf{BERT-Gen:} a woman \emph{washes} her \emph{hands} with \textbf{a \emph{sink} of \emph{soaps}}.}                                                                         \\
\multicolumn{2}{l}{\textbf{UniLM:} \emph{hands} \emph{washing} \emph{soap} in the \emph{sink}}                                                                                            \\
\multicolumn{2}{l}{\begin{tabular}[c]{@{}l@{}}\textbf{BART:} a man is \emph{washing} his \emph{hands} \textbf{in a \emph{sink} a \emph{sink}}. \end{tabular}} \\
\multicolumn{2}{l}{\textbf{T5:} \emph{hand} \emph{washed} with \emph{soap} in a \emph{sink}.}                                                                                             \\ \hline
\multicolumn{2}{l}{\textbf{Ours:} A man is \emph{washing} his \emph{hands} with \emph{soap} in a \emph{sink}.}                                                                            \\ \bottomrule
\end{tabular}
}
\caption{Sentences produced by commonly-used pretrained models given some concepts. Those sentences generated by existing pretrained models are either implausible (e.g. \emph{a sink of soaps}) or suffer from the repetition problem (e.g. \emph{in a sink a sink}), while ours generates a more natural sentence.}
\label{data_case}
\end{table}

\begin{figure*}[!t]
	\centering
	\includegraphics[width=0.9\linewidth]{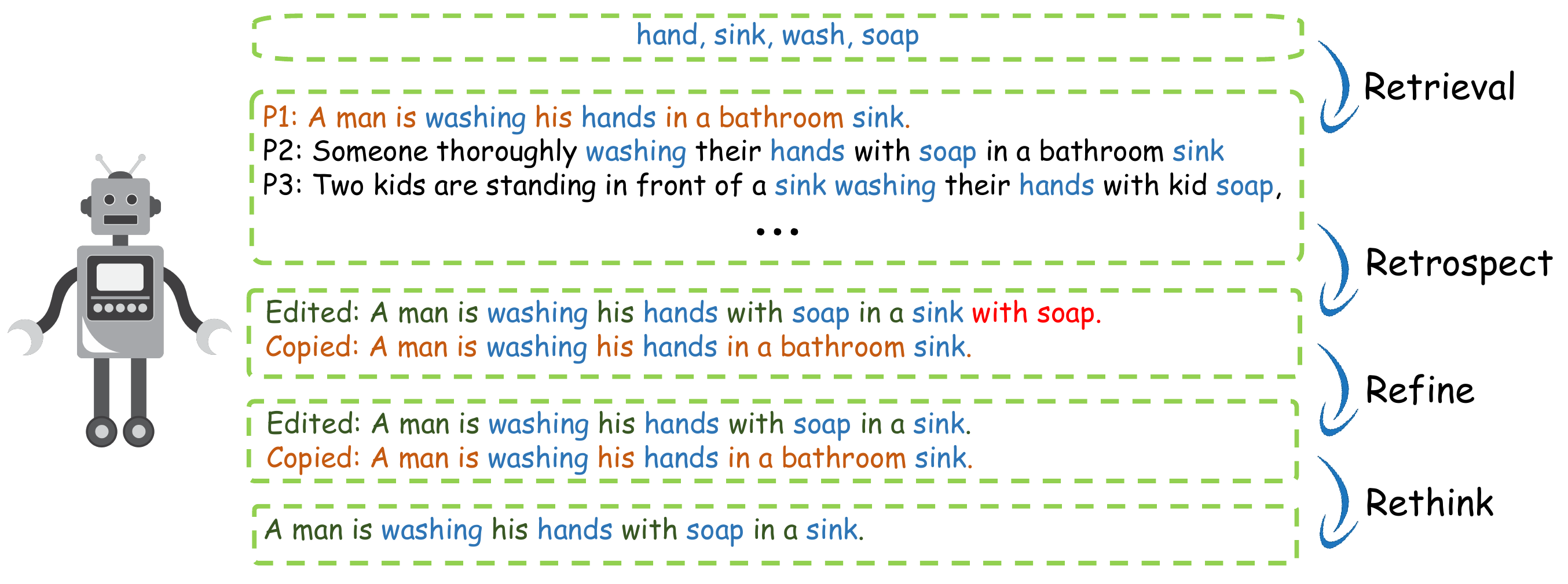}
	\caption{
		\label{fig:pipeline}
		The process of our framework generating the sentence for input concepts {\emph{hand, sink, wash, soap}}, which are marked in blue. P1, P2, P3 denote the prototypes retrieved from external corpora. The sentences followed by \emph{Edited} are the sentence generated by editing prototypes, while those followed by \emph{Copied} are copied from prototypes. The red chunk consists of repeated words and will be deleted at the refine stage. Our framework selects the better sentence as the final output at the rethink stage.
	}
	\label{fig:human_pipeline}
\end{figure*}

To strengthen the ability of generative commonsense reasoning, many researchers resort to introducing external knowledge to refine pretrained models. 
Typically, \citet{liu2021kgbart} exploit the knowledge graph ConceptNet \citep{liu2004conceptnet} to refine the BART-based model. However, knowledge graphs are usually human-annotated and might fail to cover all commonsense in daily life. Therefore, the performance of this line of work would be limited.
More recently, \citet{fan-etal-2020-enhanced} and \citet{DBLP:journals/corr/abs-2105-11174} introduce retrievers to search auxiliary information from external plain sentences, which contains enormous daily scenarios. Intuitively, the implicit commonsense knowledge within plain sentences is more abundant than that in human-annotated knowledge bases.
In this work, we extend this idea and design a novel generation framework inspired by the way of human writing.

Suppose a user is asked to write a sentence mentioning the given concepts: they may attempt to search for an ideal sentence that can be directly copied as the answer. But if the searched sentence does not mention all concepts, they may further edit it to meet the requirements, where the potential errors should be corrected. Finally, they might consider several candidate sentences through the above process and pick the most satisfying one.
Inspired by such process, in this paper, we propose a \textbf{K}nowledge-enhanced Commonsense \textbf{G}eneration framework, termed KGR$^\mathfrak{4}$. As illustrated in Figure \ref{fig:human_pipeline}, our framework consists of four steps:
1) \textbf{Retrieval}: 
We first perform retrieval to obtain prototypes for generation, where candidates are roughly retrieved by concept mapping and then a trainable scorer is used to select satisfactory candidates as prototypes.
2) \textbf{Retrospect}: A BART-based seq2seq model is employed as the generator, which exploits prototypes for better sentence generations. To encourage the generator to edit or copy prototypes as output, 
we propose \emph{retrospective training} and \emph{retrospective augmentation} to enhance the training of the generator.
3) \textbf{Refine}: 
At this stage, we train a BART-based \citep{lewis-etal-2020-bart} refiner to fix errors within generated sentences.
4) \textbf{Rethink}: Finally, we reuse the previously-trained scorer to select the best sentence from those produced by various generators.

To investigate the effectiveness of our framework, we conduct extensive experiments on the CommonGen benchmark, where experimental results and in-depth analysis demonstrate the superiority of our framework. Specifically, KGR$^\mathfrak{4}$ significantly surpasses the previous best result \citep{DBLP:journals/corr/abs-2105-11174} on the CommonGen v1.0 test set (34.40 vs. 39.70 \texttt{SPICE} points \citep{DBLP:conf/eccv/AndersonFJG16}). 
Besides, on the official test set (v1.1) of the leaderboard, KGR$^\mathfrak{4}$ achieves 33.56 \texttt{SPICE}, outperforming the previous best model by 2.48 \texttt{SPICE} points and setting a new state-of-the-art.

\begin{figure*}[!ht]
\centering
    \subfigure[Pretraining Instance]{   \label{subfig:pretrain}
    \includegraphics[width=0.38\linewidth]{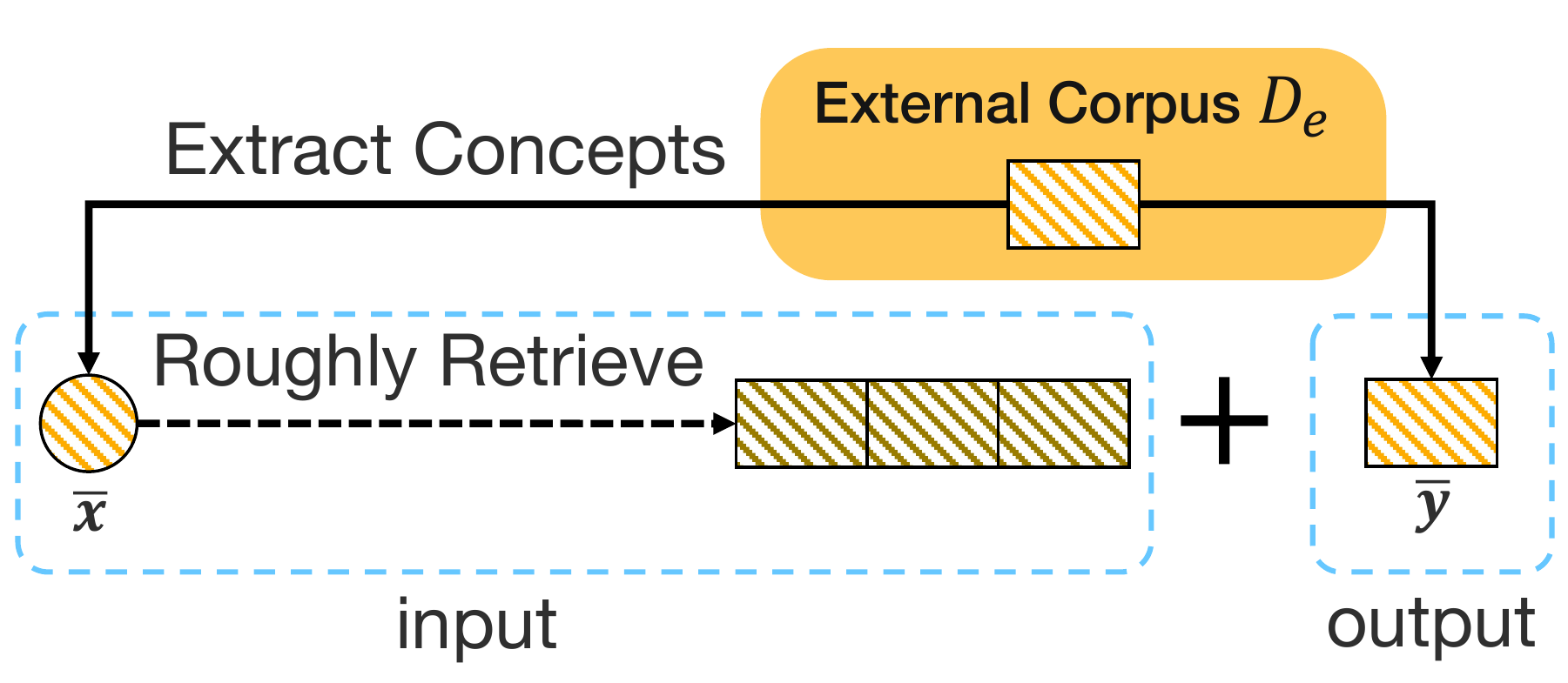}
    }
    \quad \quad
    \subfigure[Retrospective Augmented Instance]{   \label{subfig:retro_aug}
    \includegraphics[width=0.33\linewidth]{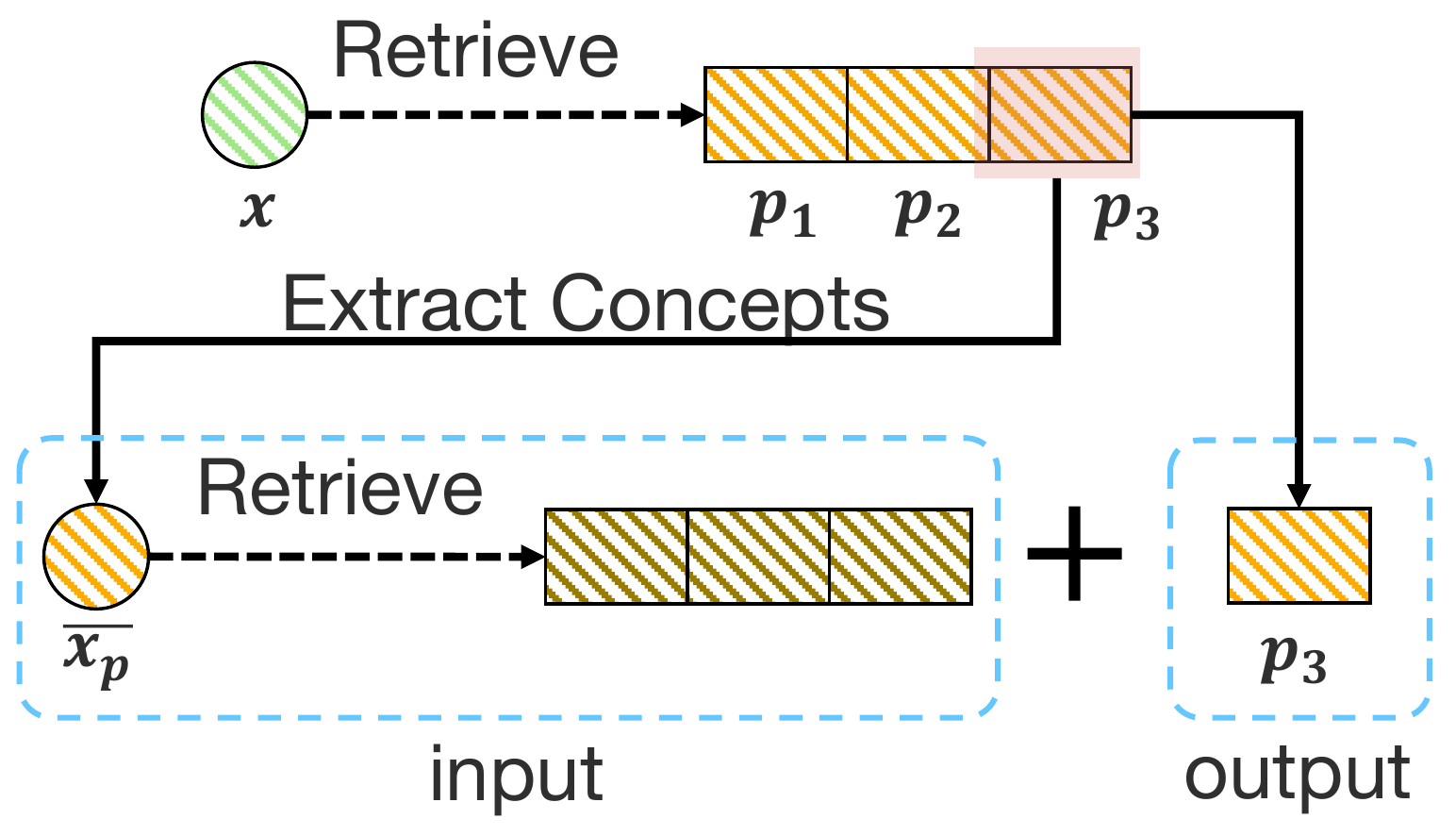}
    }
    \quad \quad
    
    \subfigure[Retrospective Training Instance (Editing)]{ \label{subfig:init}
    \includegraphics[width=0.33\linewidth]{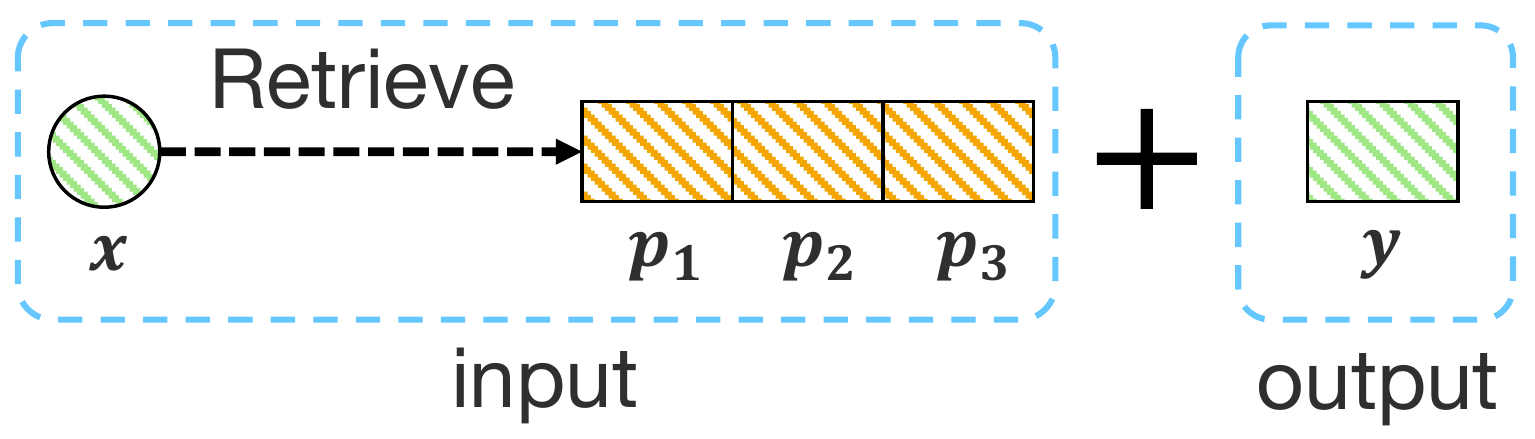}
    }
    \quad \quad
    \subfigure[Retrospective Training Instance (Copying)]{ \label{subfig:retro_train}
    \includegraphics[width=0.33\linewidth]{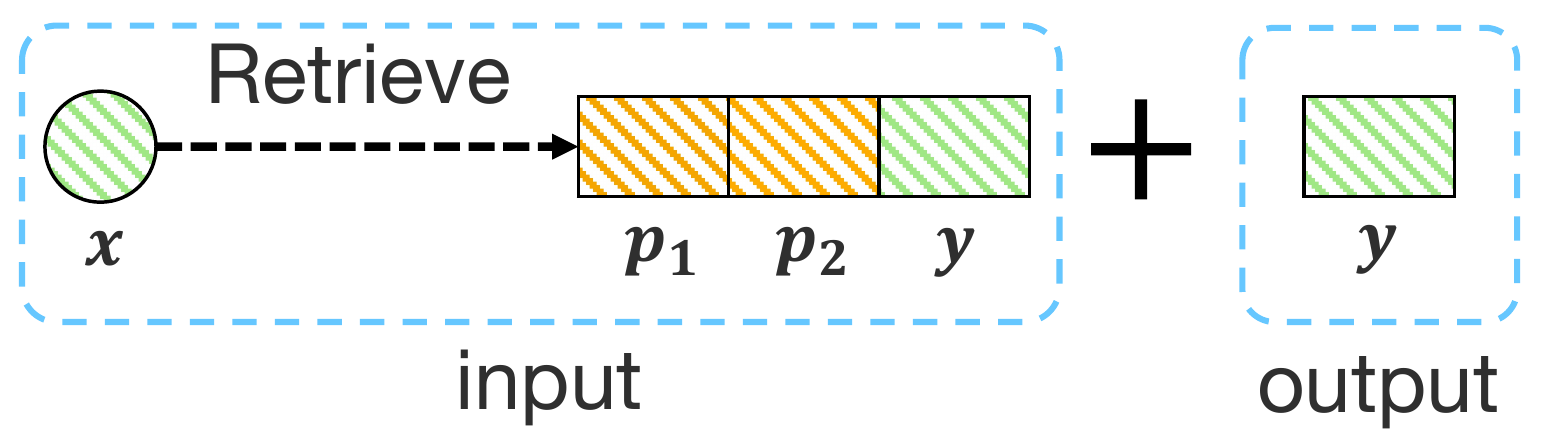}
    }
   
	\caption{
		Four types of training instances at the retrospect stage. 
		(a) A pretraining instance, of which target sentence $\bm{\Bar{y}}$ is extracted from external corpus $\mathcal{D}_e$ and input concepts $\bm{\Bar{x}}$ are extracted from $\bm{\Bar{y}}$;
		(b) A retrospective augmented instance, where both pseudo concept set $\bm{\bar{x}_p}$ and pseudo target sentence $\bm{p}_3$ are extracted from the prototype of external corpus $\mathcal{D}_e$;
		(c) A retrospective training instance encouraging the generator to edit prototypes; 
		(d) A retrospective training instance encouraging the generator to copy prototypes, where one of its prototypes (randomly chosen) is replaced with the target sentence.
	}
	\label{fig:retrospect}
\end{figure*}

\section{Related Work}
Our related work mainly includes the studies in two aspects: commonsense reasoning and utilizing commonsense knowledge in natural language generation (NLG).
\subsubsection{Commonsense Reasoning}
Recently, a series of works have been proposed to investigate the machine commonsense reasoning ability. Typically, SWAG \citep{DBLP:conf/emnlp/ZellersBSC18}, CODAH \citep{DBLP:journals/corr/abs-1904-04365}, HellaSWAG \citep{DBLP:conf/acl/ZellersHBFC19}, CommonsenseQA \citep{DBLP:conf/naacl/TalmorHLB19}, Atomic \citep{DBLP:conf/aaai/SapBABLRRSC19} are proposed to assess the commonsense reasoning ability of models for question answering. 
To enhance such ability,
\citet{DBLP:conf/acl/RajaniMXS19} collect human explanations in the form of natural language for better commonsense reasoning.
\citet{DBLP:conf/emnlp/LinCCR19} design an explainable inference framework to utilize external commonsense graphs for commonsense question answering.
Besides, \citet{DBLP:conf/emnlp/ShwartzWBBC20} study pretrained language models as an alternative of external knowledge provider to benefit commonsense question answering.
Furthermore,
since writing sentence as natural as humans is a desired ability for machines, researchers also contributed to incorporating commonsense in NLG.


\subsubsection{Utilizing Commonsense Knowledge in NLG}
To assess the machine commonsense reasoning ability in a generative manner, \citet{lin-etal-2020-commongen} propose CommonGen which asks models to generate a reasonable and fluency sentence, which mentions all the given concepts.
To enhance generative commonsense reasoning, many researchers devote to incorporating existing knowledge bases into pretrained models. 
For instance, \citet{liu2021kgbart} additionally introduce an encoder and a decoder customized to exploit the knowledge graph related to the given concepts.
\citet{zhou2021pretraining} infuse concept-centric commonsense knowledge into pretrained models via generative and contrastive objectives.
Nevertheless, the performance of these models is limited due to the low coverage and quality of used knowledge bases.
Besides, many researchers focus on utilizing implicit commonsense knowledge within plain sentences and adopt the retrieval-and-generation paradigm.
Typically,
\citet{fan-etal-2020-enhanced} equip the BART-based generation model with an enhanced knowledge injection module, which exploits the commonsense knowledge extracted from retrieved sentences. Furthermore, \citet{DBLP:journals/corr/abs-2105-11174} propose a trainable retriever to search auxiliary sentences for given concepts, and concatenate them with concepts as input for T5 \citep{DBLP:journals/jmlr/RaffelSRLNMZLL20}.

Obviously, our framework is an extension of retrieval-and-generation methods. We additionally propose three novel stages. At the retrospect stage, the generator produces sentences by copying or editing the retrieved prototypes. Generated sentences will be refined by our refiner to fix the errors at the refine stage, and the best sentence will be picked among candidate sentences produced by generators with different hyper-parameters at the rethink stage. 

\section{Our Framework}
In this section,
by simulating the process of humans writing sentences,
we propose a novel knowledge-enhanced commonsense generation framework,
which aims to generate a high-quality target sentence $\bm{y}$ containing all the concepts in set $\bm{x}=\{x_1, x_2,...,x_N\}$, where $x_i$ denotes the $i$-th concept.
As illustrated in Figure \ref{fig:human_pipeline}, our framework mainly consists of four stages: Retrieval, Retrospect, Refine and Rethink.
In the following subsections, we describe these stages in detail.





\subsection{Retrieval} \label{sec:retrieval}
Intuitively, when asked to write a sentence with given concepts, humans will always call to mind the scenarios associated with these concepts, then conceive logical sentences.
Similarly, our framework retrieves prototypes related to the given concepts from external corpora, which contain abundant scenario knowledge, then uses these prototypes as auxiliary information for the commonsense generation.


We firstly use the concept mapping~\cite{DBLP:journals/corr/abs-2105-11174} and roughly search candidate prototypes for the concept set $\bm{x}$ from an external corpus $\mathcal{D}_e$, which consists of caption sentences describing the daily scenario. Then, we train a RoBERTa-based binary classifier as a scorer to semantically evaluate candidate prototypes.
The scorer takes the concatenation of $\bm{x}$ and a candidate prototype $\bm{p'}$ as input and outputs a score $f_s(\bm{x}, \bm{p'})$ indicating whether $\bm{p'}$ is related to $\bm{x}$.
To optimize the scorer, we construct a temporary training set from $\mathcal{D}_e$ and the CommonGen training set $\mathcal{D}$, which consists of $\langle \bm{x}, \bm{y} \rangle$. 
More specifically, for each concept set $\bm{x}$ from $\mathcal{D}$, we take its corresponding sentence $\bm{y}$ to construct a positive sample and randomly select sentences from $\mathcal{D}_e$ to produce negative samples.
Finally, we apply this scorer to semantically evaluate candidate prototypes for each concept set.
As implemented in \citet{DBLP:journals/corr/abs-2105-11174}, we select the top-3 scored candidate prototypes as the final prototype set.

\subsection{Retrospect} \label{sec:retrospect}

We then construct a generator to exploit prototypes for sentence generations.
Note that given the prototypes, 
humans might glance at these prototypes before sentence writing. 
If one of these prototypes is good enough, 
they could directly copy it as output. 
Otherwise, they might write the sentence by editing prototypes. Thus, we believe both editing and copying prototypes are crucial for generating high-quality sentences.
For that purpose, we propose two novel strategies to enhance the training of the generator.


\subsubsection{Generator}
Following previous studies~\cite{liu2021kgbart,fan-etal-2020-enhanced}, 
we adopt BART-large \cite{lewis-etal-2020-bart} to establish our generator,
which takes the concept set and retrieved prototypes as input and generates a target sentence mentioning all concepts.
To make the model fully absorb the commonsense knowledge from the external corpus $\mathcal{D}_e$, we pretrain the generator using a large-scale pretraining dataset $\mathcal{D}_{pt}$ constructed from $\mathcal{D}_e$.
In specific, the construction steps of $\mathcal{D}$ are as follows:
taking the sentence $\bm{\Bar{y}}$ in $\mathcal{D}_e$ as the target sentence, we use spaCy \citep{ines_montani_2021_4900666} to perform POS tagging on $\bm{\Bar{y}}$, and sample several lemmatized Verbs, Nouns, and Proper Nouns as the pseudo concepts $\Bar{\bm{x}}$.
We pair $\Bar{\bm{x}}$ with $\bm{\Bar{y}}$ to form a pretraining instance (Figure \ref{subfig:pretrain}).
Considering the training efficiency of the generator, we only roughly retrieve the prototypes via concept matching during pretraining. Formally, we use $\mathcal{D}_{pt}$ to pretrain our generator in the following way:

\begin{equation}
    \mathcal{L}^{pt}_{G}= -\sum_{t=1}^{|\bm{\Bar{y}}|}\log p(\Bar{y}_t|\bm{\Bar{x}},\{\bm{\Bar{p}_1},\bm{\Bar{p}_2},\bm{\Bar{p}_3}\},\bm{\Bar{y}}_{<t}),
\end{equation}
where $\bm{\Bar{p}_*}$ is the roughly retrieved prototypes, and $\Bar{y}_t$ denotes the $t$-th token of $\bm{\Bar{y}}$. 

Afterwards, we propose two novel strategies to finetune the generator so as to enhance the training of the generator.
Since the retrieved prototypes related to the given concepts are usually human-created, they are natural and coherent.
We argue that letting the generator learn more about how to generate such prototypes will help the generator better capture the meaning of these concepts, and further benefits the subsequent commonsense generation.
Thus, we propose \textbf{retrospective augmentation} strategy to expand $\mathcal{D}$ into a new training set $\mathcal{D}_{ft}$.
As shown in Figure \ref{subfig:retro_aug},
following the concepts extracting process mentioned before, 
we extract pseudo concepts $\Bar{\bm{x}}_p$ from the prototype $\bm{p}_3$, and pair $\Bar{\bm{x}}_p$ with $\bm{p}_3$ to form a retrospective augmented instance, which will be added into $\mathcal{D}_{ft}$.


Most importantly, we propose \textbf{retrospective training} strategy to enhance the editing and copying ability of our generator.
Unlike the previous pretraining only based on the MLE training objective, our training objective involves two loss items.
Formally, given a training instance $\langle \bm{x}, \bm{y} \rangle$ from $\mathcal{D}_{ft}$ and its retrieved prototypes $\{\bm{p_1},\bm{p_2},\bm{p_3}\}$, we define the joint training objective as follows:

\begin{equation}
    \mathcal{L}^{ft}_{G} = (1-\lambda)\mathcal{L}_{edit} + \lambda \mathcal{L}_{copy}, \label{eq:lambda_obj}
\end{equation}
\begin{equation}
    \mathcal{L}_{edit}= -\sum_{t=1}^{|\bm{y}|}\log p(y_t|\bm{x},\{\bm{p_1},\bm{p_2},\bm{p_3}\},\bm{y}_{<t}),
\end{equation}
\begin{equation}
    \mathcal{L}_{copy}= -\sum_{t=1}^{|\bm{y}|}\log p(y_t|\bm{x},\{\bm{p_1},\bm{p_2},\bm{y}\},\bm{y}_{<t}),
\end{equation}
where $\lambda$ is the hyper-parameter to balance $\mathcal{L}_{edit}$ and $\mathcal{L}_{copy}$.
$\mathcal{L}_{edit}$ denotes the editing training objective. As illustrated in Figure \ref{subfig:init}, the generator generates the target sentence $\bm{y}$ by editing the retrieved prototypes $\{\bm{p_1},\bm{p_2},\bm{p_3}\}$.
$\mathcal{L}_{copy}$ is the copying training objective. We randomly replace one of the prototypes in $\{\bm{p_1},\bm{p_2},\bm{p_3}\}$ with the target sentence $\bm{y}$, which is illustrated in Figure \ref{subfig:retro_train}. 
By using $\mathcal{L}_{copy}$, 
we expect that the generator is able to detect the high-quality prototype and copy it as output.

\subsection{Refine} \label{sec:refine}
Similar to previous works \citep{lewis-etal-2020-bart, fan-etal-2020-enhanced} focusing on CommonGen, we find that our generator suffers from the \emph{degeneration problem} \citep{DBLP:conf/iclr/WelleckKRDCW20}. 
To ensure these sentences are grammatically correct, we propose a refiner to fix potential errors within generated sentences.


\subsubsection{Refiner}
To better model our refiner,
we first analyze the errors within the generated sentences,
which can be classified into the following two types:
1) \emph{Repetition error}. The generator sometimes pays too much attention to what it has recently produced \citep{DBLP:conf/acl/LewisDF18}, and thus it tends to generate similar text multiple times. 
For example, ``\emph{in a sink a sink}'' shown in Table \ref{data_case};
2) \emph{Misspelling}. Since the BART-based generator tokenizes the sentences using the same byte-pair encoding as GPT-2 \citep{radford2019language}, some characters and spaces in sentence should be predicted, while the generator occasionally misses them, leading to misspelling words (i.e., ``\emph{wash hands}'' is incorrectly output as ``\emph{wsh hands}'' and ``\emph{washhands}'').

To deal with the errors mentioned above, we construct a BART-based refiner, which takes the the candidate target sentence generated by our generator as input and outputs the corrected target sentence.
To this end, we construct training instances based on $\mathcal{D}_e$.
Given the sentence $\bm{\Bar{y}}$ from $\mathcal{D}_e$, we generate a perturbed sentence $\bm{\hat{y}}$ by incorporating above-mentioned errors into $\bm{\Bar{y}}$:
either randomly repeat a word sequence of $\bm{\Bar{y}}$ to simulate repetition errors, 
or randomly remove characters or spaces in $\bm{\Bar{y}}$ to introduce the misspelling words.
Then, we require the refiner to recover $\bm{\Bar{y}}$ from $\bm{\hat{y}}$ via the following auto-encoding training objective:
\begin{equation}
    \mathcal{L}_R = - \log p(\bm{\Bar{y}}|\bm{\hat{y}}) = -\sum_{t=1}^{|\bm{\Bar{y}}|} \log p(\Bar{y}_t|\bm{\hat{y}},\bm{\Bar{y}}_{<t}).
    \label{for:autoencoder}
\end{equation}
In this way, our refiner is trained to correct potential errors within candidate sentences, especially repetition and misspelling errors.


\subsection{Rethink}


The above processes describe in detail how to generate a target sentence for a given concept set.
However, this process is not entirely consistent with how humans write sentences: humans usually produce multiple sentences in different ways (copying or writing) and then select the best one from them.
To further improve quality of the generated sentences, we propose to employ the generator with different $\lambda$s to generate multiple target sentences, all of which are then refined by our proposed refiner.
Afterwards, we reuse the previously-trained scorer (See Section \ref{sec:retrieval}) to semantically evaluate these target sentences.
During this process, we feed the concatenation of the given concept set and the refined sentence to the scorer, and select the final outputted sentence with the highest score.

\section{Experiments}
\subsection{Settings} \label{sec:exp_settings}
\subsubsection{Dataset and Metrics}
Following previous studies,
we use the CommonGen dataset constructed by \citet{lin-etal-2020-commongen}.
We show the basic statistics of this dataset in Table \ref{dataset_statistics}. 
Note that all concept-pairs and concept-triples of test set are unseen in the training set, which requires models to generalize well on the unseen combinations of concepts.

As described above, our framework involves an external corpus, which serves three purposes: 1) Being the retrieval pool where the prototypes come (See Section 3.1); 2) Constructing pseudo
instances for retrospective augmentation and pretraining (See Section 3.2); 3) Establishing training and validation sets to train the refiner (See Section 3.3).
We construct this corpus by combining 3M image and video captions of several datasets: Activity \citep{Krishna_2017_ICCV}, MultiNLI \citep{DBLP:conf/naacl/WilliamsNB18}, SNLI \citep{DBLP:conf/emnlp/BowmanAPM15}, Vatex \citep{DBLP:conf/iccv/WangWCLWW19}, MSCOCO \citep{DBLP:conf/eccv/LinMBHPRDZ14} and \citep{DBLP:journals/tacl/YoungLHH14}.


\renewcommand\arraystretch{1.0}
\begin{table}[]
\small
\centering
\scalebox{0.95}{
\begin{tabular}{l|ccc}
\bottomrule
\textbf{Statistics}             & \textbf{Train} & \textbf{Validation} & \textbf{Test} \\ \hline
\textbf{\#Concept Sets}           & 32,651         & 993          & 1,497         \\
\ \ \ -Size = 3                       & 25,020         & 493          & -             \\
\ \ \ -Size = 4                       & 4,240          & 250          & 747           \\
\ \ \ -Size = 5                       & 3,391          & 250          & 750           \\ \hline
\textbf{\#Sentences}              & 67,389         & 4,018        & 7,644         \\ \hline
\textbf{\ \ Unseen Concepts}        & -              & 6.53\%       & 8.97\%        \\
\textbf{\ \ Unseen Concept-Pairs}   & -              & 96.31\%      & 100.00\%      \\
\textbf{\ \ Unseen Concept-Triples} & -              & 99.60\%      & 100.00\%      \\ \bottomrule
\end{tabular}
}
\caption{The basic statistics of the CommonGen dataset. \#Concept is the number of concepts each concept set contains. Unseen concept compositions (i.e., concept, concept-pair, concept-triple) denote the ratio of unseen compositions in the training set, which propose challenges on the generalization of models.}
\label{dataset_statistics}
\end{table}

Following \citet{lin-etal-2020-commongen}, we use \texttt{BLEU-4} \citep{DBLP:conf/acl/PapineniRWZ02}, \texttt{CIDEr} \citep{DBLP:conf/cvpr/VedantamZP15}, \texttt{SPICE} \citep{DBLP:conf/eccv/AndersonFJG16} as our evaluation metrics. Since \citet{lin-etal-2020-commongen} claim that \texttt{SPICE} is most relevant to human evaluation, we use it as our prior metric.

\subsubsection{Baselines}
We compare KGR$^\mathfrak{4}$ with several competitive generation models: 
\begin{itemize}
    \item \textbf{EKI-BART} \citep{fan-etal-2020-enhanced}. It is a knowledge-enhanced model based on vanilla BART, which retrieves a prototype for better sentence generation;
    \item \textbf{KG-BART} \citep{liu2021kgbart}. It pretrains BART via a masked concept prediction task, and further leverages the knowledge graph to enhance the encoder and decoder;
    \item \textbf{CLAM} \citep{zhou2021pretraining}. It introduces generative and contrastive objectives into pretrained text generation models, so as to better learn concept-centric commonsense knowledge;
    \item \textbf{RE-T5} \citep{DBLP:journals/corr/abs-2105-11174}. It is a T5-based model equipped with a trainable retriever to retrieve prototypes as the auxiliary input.
\end{itemize}

Moreover, we report the performance of several commonly-used pretrained generation models,including: \textbf{GPT-2} \citep{radford2019language}, \textbf{BERT-Gen} \citep{pmlr-v119-bao20a}, \textbf{UniLM} \citep{NEURIPS2019_c20bb2d9}, \textbf{BART} \citep{lewis-etal-2020-bart}, \textbf{T5-base} \citep{JMLR:v21:20-074}, and \textbf{T5-large} \citep{JMLR:v21:20-074}. 
Except T5-base and T5-large that add a prompt into the beginning of input sequence,
all other models take the combination of concepts as input and output the description sentence.


\subsubsection{Implementation Details}
At the retrieval stage, we select 3 negative samples for each positive sample. We optimize the RoBERTa-based scorer using the Adam optimizer \citep{DBLP:journals/corr/KingmaB14} with a learning rate of 2e-5 for 3 epochs, and set the batch-size to 32. 
At the retrospect stage, we pretrain the generator for 80,000 steps using the pseudo instances constructed from the external corpus and then finetune the model parameters for 2,000 steps, where the learning rate of the Adam optimizer is set as 2e-5, and the batch size is 16. 
In both pretraining and retrospective augmentation, we sample 5 concepts from each sentence.
At the refine stage, we construct the training and validation set for the refiner from the external corpus. In both training and validation sets, 5\% of instances are sampled to produce the perturbed sentences, 50\% of the perturbed sentences contain misspelling errors, while the others contain repetition errors. During this process,
we remove 1\% characters and 10\% spaces from the instances containing misspelling errors,
and repeat sentence segments to construct instances with repetition errors.
We update the parameters of refiner for 2,000 steps and keep the rest hyper-parameters same as the generator.
Particularly, we employ early-stopping when training scorer, generator, and refiner.

\renewcommand\arraystretch{1.0}
\begin{table*}[!htbp]
\centering
\begin{tabular}{lcccc}
\bottomrule
Model       & BLEU-4(lb) & CIDEr(lb)  & \textbf{SPICE(lb)} & SPICE(v1.0) \\ \hline
GPT-2~\citep{radford2019language}       & 26.833 & 12.187 & 23.567         & 25.90\\
BERT-Gen~\citep{pmlr-v119-bao20a}    & 23.468 & 12.606 & 24.822         & 27.30\\
UniLM~\citep{NEURIPS2019_c20bb2d9}       & 30.616 & 14.889 & 27.429         & 30.20\\
BART~\citep{lewis-etal-2020-bart}        & 31.827 & 13.976 & 27.995         & 30.60\\
T5-base~\citep{JMLR:v21:20-074}     & 18.546 & \ \ 9.399  & 19.871         & 22.00\\
T5-large~\citep{JMLR:v21:20-074}    & 31.962 & 15.128 & 28.855         & 31.60\\ \hline
EKI-BART~\citep{fan-etal-2020-enhanced}    & 35.945 & 16.999 & 29.583         & 32.40\\
KG-BART~\citep{liu2021kgbart}     & 33.867 & 16.927 & 29.634         & 32.70\\ 
CALM(T5-base)~\citep{zhou2021pretraining}        & -      & -      & -              & 33.00\\
RE-T5~\citep{DBLP:journals/corr/abs-2105-11174}  & 40.863 & 17.663 & 31.079 & 34.30       
\\
\hline
KGR\textsuperscript{4}  & \textbf{42.818} & \textbf{18.423} & \textbf{33.564} & \textbf{39.70} \\
\bottomrule
\end{tabular}
\caption{\label{results2}
Experimental results on the CommonGen benchmark. 
*(lb) means the results are shown on the official leaderboard.
*(v1.0) indicates the evaluation using the old evaluation protocol. Please note that \texttt{SPICE} is our most important metric.}
\label{main_res}
\end{table*}

\subsection{Overall Results}

Table \ref{main_res} lists the overall results of various models. We can observe that KGR$^\mathfrak{4}$ performs best among all models. Compared with the previous best model RE-T5, KGR$^\mathfrak{4}$ surpasses it by 1.955 \texttt{BLEU-4} and 0.760 \texttt{CIDEr} points. Meanwhile, KGR$^\mathfrak{4}$ achieves the highest \texttt{SPICE} point 33.564, setting a new SOTA on the official leaderboard\footnote{\url{https://inklab.usc.edu/CommonGen/leaderboard.html}}.
Please note that both architectures of EKI-BART and KG-BART are modified to exploit external knowledge,
while our framework is completely data-driven, independent on specific model architecture.
Thus, our framework can be directly applied to any task-oriented models, such as EKI-BART and KG-BART, to gain further improvements.

\begin{figure}[!t]
	\centering
	\includegraphics[width=0.85\linewidth]{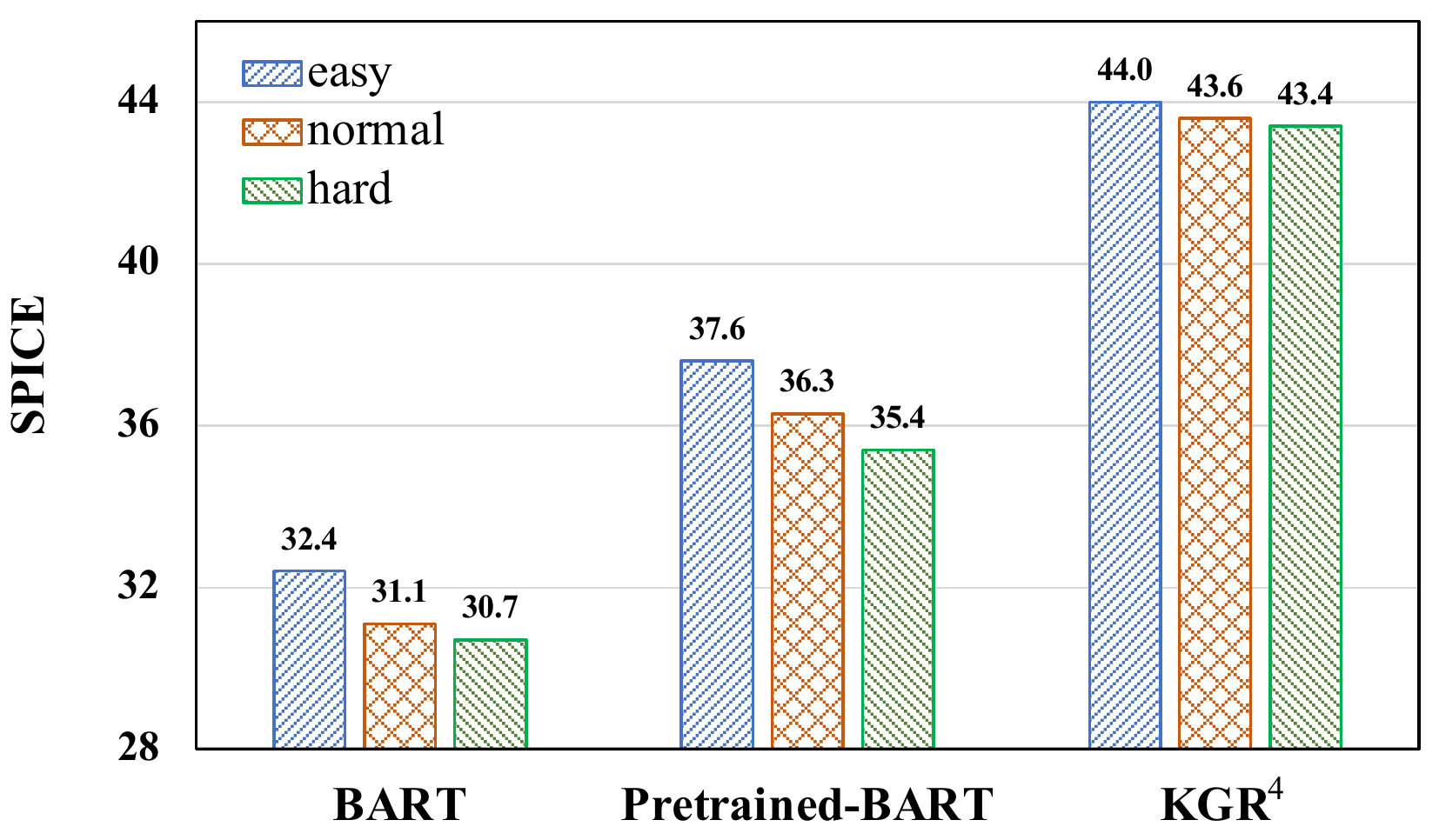}
	\caption{
		\texttt{SPICE} points on test sets with various difficulties.
	}
	\label{fig:multi_hops}
\end{figure}

\subsubsection{Performance on Instances with Various Difficulties} \label{sec:exp_difficulties}
Following \citet{lin-etal-2020-commongen}, we extract 5-size concept-sets (each concept-set consists of five concepts) from the test set, and classify them into 3 categories: \textbf{easy}, \textbf{normal}, \textbf{hard}, representing difficulties of generating sentences about them.
Here we estimate the difficulty of each concept-set based on the following fact:
if there are more concepts directly connected on ConceptNet, then it is easier to write sentences about them. For each 5-size concept-set, there are 10 concept pairs (pick two among five concepts). We count the one-hop connections of concept pairs that are directly connected on ConceptNet, and then empirically divide concept-sets into \textbf{hard}\textbar\textbf{normal}\textbar\textbf{easy} categories if they contain [0,2]\textbar[3,5]\textbar[6,10] one-hop connections.

We illustrate the performance of BART, BART pretrained using the external corpus (Pretrained-BART), and KGR$^\mathfrak{4}$ on the test sets with various difficulties in Figure \ref{fig:multi_hops}.
We can find that our framework gains the best results on all test sets, strongly demonstrating its superiority. Besides, it is interesting to observe that our framework reaches the lowest performance decline when switching the test set from the easy to the hard one. This indicates our framework can better deal with all the given concepts regardless of their difficulties.


\renewcommand\arraystretch{1.0}
\begin{table}[t]
\centering
\begin{tabular}{lc}
\bottomrule
\multicolumn{1}{l}{model} &   SPICE(v1.0) \\ \hline
BART                      & 30.60          \\
\ \ +pretraining             & 33.10         \\
\ \ +retrieval                & 36.60          \\
\ \ +retrospective training   & 38.30          \\
\ \ +retrospective augmentation        & 39.20          \\
\ \ +refine                   & 39.40          \\
\ \ +rethink                  & \textbf{39.70} \\ \bottomrule
\end{tabular}
\caption{Ablation study of KGR$^\mathfrak{4}$. }
\label{tab:ablation}
\end{table}

\subsection{Analysis}

We further conduct in-depth analyses
to investigate the following problems: 
\textbf{Q1}: Do all strategies and stages of KGR$^\mathfrak{4}$ take effects? 
\textbf{Q2}: How the hyper-parameter $\lambda$ affects the generator?
\textbf{Q3}: Can the refine stage alleviate repetition and misspelling errors?

\subsubsection{Ablation Study} \label{sec:ablation_study}
To explore the effectiveness of different stages and strategies, we report the performance of different variants of our framework.
\begin{itemize}
    \item \textbf{BART}. It is a BART-large model directly supervised by the CommonGen training set.
    \item \textbf{+pretraining}. It is the variant firstly pretrained using the pretraining instances constructed from the external corpus, and then finetuned using the CommonGen training set. 
    \item \textbf{+retrieval}. To construct this variant, we further apply retrieval during both pretraining and finetuning for the generator (See Section \ref{sec:retrieval}).
    \item \textbf{+retrospective training}. In this variant, the pretrained generator is finetuned via our proposed retrospective training strategy using the CommonGen training set (See Section \ref{sec:retrospect}).
    \item \textbf{+retrospective augmentation}. The generator of this variant is further enhanced with retrospective augmentation strategy.
    \item \textbf{+refine}. It means the generated sentences are further refined through the refine stage. 
    \item \textbf{+rethink}. indicates the output sentences are selected among predicted sentences of models using various $\lambda$s.
\end{itemize}

From Table \ref{tab:ablation}, we can observe that all our stages and strategies take effects in our framework. 
Particularly, when sequentially applying our retrospective training and retrospective augmentation strategies, our framework achieves 1.70 and 0.90 \texttt{SPICE} point improvements over its previous version, respectively. These results strongly demonstrate the effectiveness of these two strategies.



\begin{figure}[!t]
	\centering
	\includegraphics[width=0.87\linewidth]{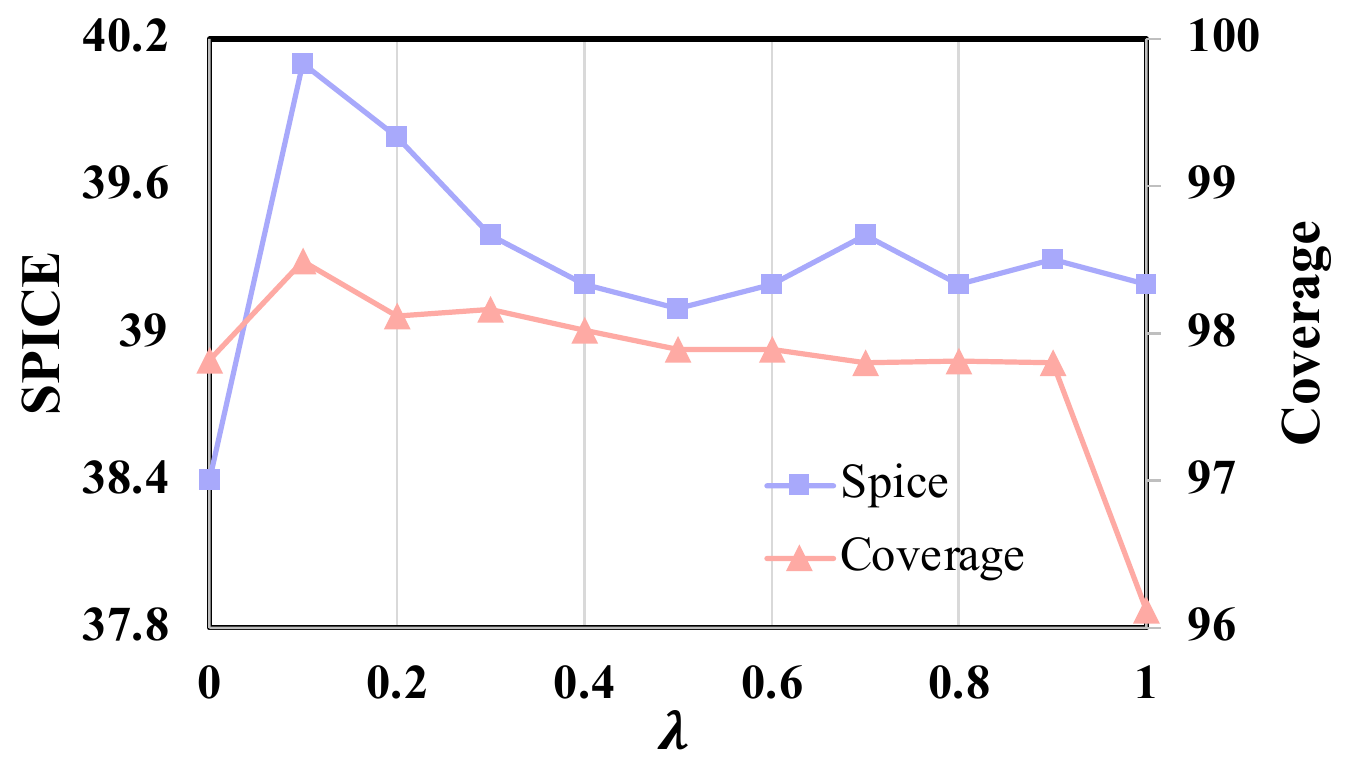}
	\caption{
		\label{fig:spice_coverage}
		\texttt{SPICE}(v1.0) and \texttt{Coverage} points computed on CommonGen test set with various $\lambda$s.
	}
\end{figure}

\subsubsection{Effect of Hyper-parameter $\lambda$} \label{sec:exp_retrospect}
As described in Section \ref{sec:retrospect},
our generator introduces an important hyper-parameter $\lambda$ to control the joint training objective listed ( See Section \ref{eq:lambda_obj}),
which has a crucial effect on the performance of our framework.
Thus, we vary $\lambda$ from 0 to 1 with an increment of 0.1 at each step, and inspect the performance of our framework using different $\lambda$s. 
In addition to our prior metric \texttt{SPICE}, we introduce another metric \texttt{Coverage} that is defined as the average percentage of input concepts occurring within output target sentences.

As shown in Figure \ref{fig:spice_coverage},
our framework always achieves satisfactory performance with any $\lambda$ in terms of \texttt{SPICE} and \texttt{Coverage}.
Note that when $\lambda$ is set to 0, our framework obtains 38.4 \texttt{SPICE} points, which is significantly inferior to those of other settings, indicating that editing retrieved prototypes regardless of copying may fail to produce high-quality sentences.
Meanwhile, we find that the \texttt{Coverage} value of our framework drops sharply when $\lambda$ increases from 0.9 to 1. The underlying reason is that the generator with $\lambda=1$ tends to copy rather than edit prototypes, while the copied one may not mention all the given concepts, leading to the lower \texttt{coverage}. 
Particularly, when setting $\lambda$ to 0.1,
our framework reaches the highest \texttt{SPICE} and \texttt{Coverage} values, demonstrating the effectiveness of our retrospect stage. 

\begin{figure}[!t]
	\includegraphics[width=0.85\linewidth]{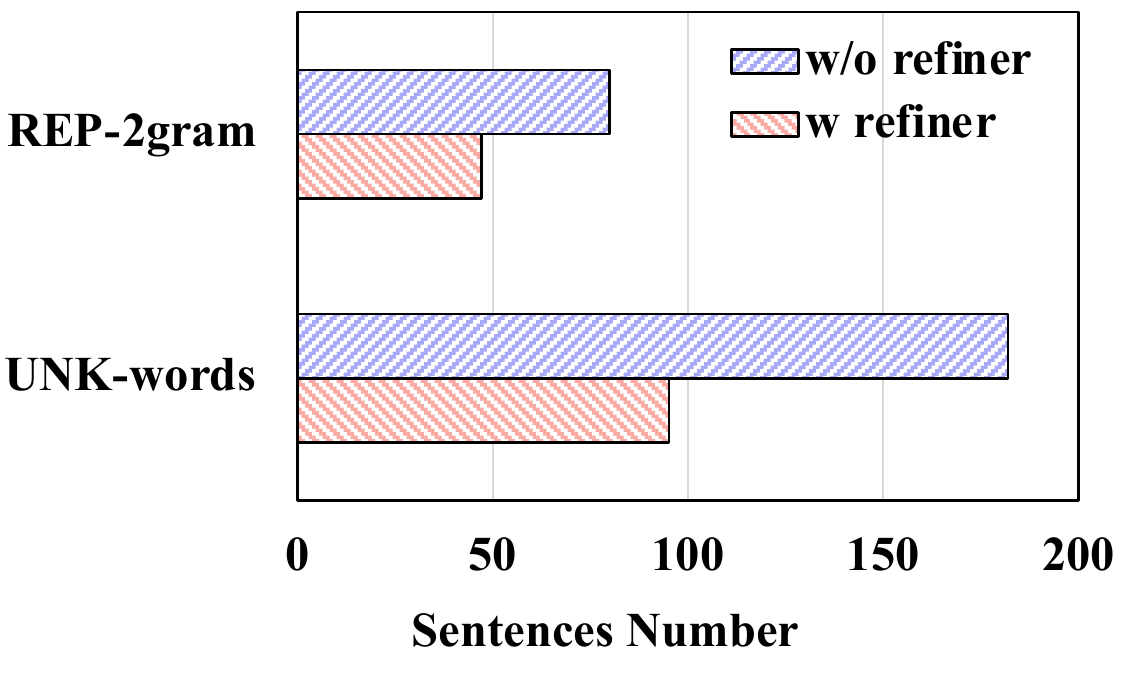}
	\caption{
		\textbf{REP-2gram}, and \textbf{UNK-words} without (w/o) and with (w) our refiner. Our refiner significantly reduces the numbers of sentences containing two types of errors. w/o refiner is the variant \emph{+retrospective augmentation} mentioned in ablation study.
	}
	\label{tab:refine}
\end{figure}

\renewcommand\arraystretch{1.2}
\begin{table}[t]
\scalebox{0.86}{
\begin{tabular}{l}
\bottomrule
\textbf{w/o refiner:}                                          \\
A \emph{dog} \emph{splashes} through a \emph{puddle} of \emph{water} \underline{in a \emph{puddle}} in
the \emph{rain} . \\
\textbf{w refiner:}                                            \\
A \emph{dog} \emph{splashes} through a \emph{puddle} of \emph{water} in the \emph{rain} .             \\ \hline
\textbf{w/o refiner:}                                          \\
Bearded \underline{manin} white shirt \emph{demonstrates} \emph{steps} to \emph{tying}
\emph{necktie}.     \\
\textbf{w refiner:}                                            \\
Bearded man in white shirt \emph{demonstrates} \emph{steps} to \emph{tying}
\emph{necktie}.    \\ \bottomrule
\end{tabular}
}
\caption{Two examples of sentences produced by our framework without and with the refine stage. Concepts appealing in sentences are marked in \emph{italics}. Erroneous words are underlined.}
\label{case_study}
\end{table}

\subsubsection{Impact of the Refine Stage} \label{sec:exp_refine}
To quantify impact of our refiner,
we conduct statistics on the two kinds of errors occurring within the sentences generated by our refiner.
Concretely, we count the number of sentences containing repeated $n$-grams to measure repetition errors (\textbf{REP-ngram}). 
Besides,
we roughly consider the output words, which are unseen in both the CommonGen dataset and the external corpus, as misspelling ones. We count the number of sentences containing those words (\textbf{UNK-words}).
The results of two metrics without and with our refine stage are shown in Figure \ref{tab:refine}. Obviously, through refine stage, both types of errors are significantly reduce. Specifically, 40\% of the repetition and 50\% of the misspelling errors are corrected at this stage, strongly demonstrating the effectiveness of our refiner.

Table \ref{case_study} lists two examples of the sentences without and with our refiner. In the first example, we can find that the phrase ``\emph{a puddle}'' is repeated before the refine stage. By contrast, our refiner detects this error and removes the repeated phrase as well as the word ``\emph{in}'', generating a more plausible sentence.
In the second example, the space between ``\emph{man}'' and ``\emph{in}'' is missed, while our refiner not only adds it but also keeps the rest of the sentence unchanged.

\section{Conclusion}
In this paper, we have proposed KGR$^\mathfrak{4}$, a commonsense generation framework. Our framework mainly separates four key stages, i.e., retrieval, retrospect, refine, and rethink, to imitate corresponding human behaviors in writing. On the commonly-used datasets, KGR$^\mathfrak{4}$ outperforms several competitive baselines, setting a new state of the art on the official board.
Besides, our study suggests following points: 1) Each proposed component in our framework has its unique function in commonsense generation, and worth to be further explored; 2) The retrieval and retrospect stages play more crucial roles on improving the commonsense reasoning ability; and 3) The refine and rethink stages enable our framework to proofread the generated candidates. 
In the future, we plan to explore variational models \citep{DBLP:conf/emnlp/ZhangXSDZ16, DBLP:conf/emnlp/ZhangXSLJDZ16, DBLP:conf/aaai/SuWXLHZ18, DBLP:journals/isci/SuWZWQX18} to refine our framework. Besides, we will generalize our framework to other conditional text generation tasks.

\section*{Acknowledgments}
The project was supported by 
National Natural Science Foundation of China (No. 62036004, No. 61672440),  
Natural Science Foundation of Fujian Province of China (No. 2020J06001),
Youth Innovation Fund of Xiamen (No. 3502Z20206059),
National Key Research and Development Program of China (No. 2018YFB1403202),
and Alibaba Group through Alibaba Innovative Research Program.
We also thank the reviewers for their insightful comments.

\bibliography{aaai22}

\end{document}